\batchmode
\makeatletter
\def\input@path{{/home/macer/Documents/paper//}}
\makeatother
\documentclass[10pt,letterpaper,twocolumn]{article}
\usepackage[T1]{fontenc}
\usepackage{array}
\usepackage{graphicx}
\usepackage[unicode=true,
 bookmarks=true,bookmarksnumbered=true,bookmarksopen=true,bookmarksopenlevel=1,
 breaklinks=false,pdfborder={0 0 0},backref=false,colorlinks=false]
 {hyperref}
\hypersetup{pdftitle={Your Title},
 pdfauthor={Your Name},
 pdfpagelayout=OneColumn, pdfnewwindow=true, pdfstartview=XYZ, plainpages=false}

\makeatletter

\pdfpageheight\paperheight
\pdfpagewidth\paperwidth

\providecommand{\tabularnewline}{\\}

\@ifundefined{date}{}{\date{}}
\usepackage[caption=false,font=footnotesize]{subfig}
\usepackage{wacv}
\usepackage{times}
\usepackage{epsfig}
\usepackage{graphicx}
\usepackage{amsmath}
\usepackage{amssymb}

\@ifundefined{showcaptionsetup}{}{%
 \PassOptionsToPackage{caption=false}{subfig}}
\usepackage{subfig}
\makeatother

\begin{document}
\wacvfinalcopy
\def\wacvPaperID{} 

\title{Fast Vehicle Detection in Aerial Imagery
\author{Jennifer Carlet \\ KeyW Corp.\\ {\small Beavercreek, OH} \and Bernard Abayowa \\ Sensors Directorate, Air Force Research Lab\\ {\small WPAFB, OH} }}
\maketitle
\begin{abstract}
\textit{In recent years, several real-time or near real-time object
detectors have been developed. However these object detectors are
typically designed for first-person view images where the subject
is large in the image and do not directly apply well to detecting
vehicles in aerial imagery. Though some detectors have been developed
for aerial imagery, these are either slow or do not handle multi-scale
imagery very well. Here the popular YOLOv2 detector is modified to
vastly improve it's performance on aerial data. The modified detector
is compared to Faster RCNN on several aerial imagery datasets. The
proposed detector gives near state of the art performance at more
than 4x the speed.} 
\end{abstract}

\section{Introduction}

Object detection from ground view is a popular problem with a lot
of interest from the academic computer vision community. Detection
from aerial views, while there is some interest, is significantly
less studied. Consequently recent advancements are primarily large
in image object detection and classification, mostly using deep convolutional
neural networks (CNNs). Often these neural networks do not work well
when directly applied to small in image objects. However the networks
can often be modified to improve their performance on this type of
data. 

Of particular interest is detecting vehicles from aerial platforms
at near real-time speeds. While Faster RCNN\cite{renNIPS15fasterrcnn}
has been proven to be effective at detecting vehicles in aerial imagery,
it is unable to reach anywhere near the real time speeds desired for
many applications. Other methods use sliding window techniques which
can also be slow. Newer detectors that can run on whole images are
much faster but have yet to be proven on aerial imagery. In this paper,
an open-source fast deep CNN is modified into a near real-time multi-scale
detector for aerial imagery.

The rest of this paper is organized as follows. Section 2 introduces
the deep CNNs used in this paper. Section 3 covers the aerial imagery
datasets used. How the net was modified is described in Section 4.
Section 5 gives the results. The paper is concluded in Section 6.

\section{Deep learning object detection algorithms}

Since a deep CNN easily won the 2012 ImageNet competition, CNNs have
become the state of the art in object detection in images. Whereas
before hand crafted features such gradients, color, etc., were used
to detect objects; now CNNs can automatically learn which features
are relevant for detection. This work focused on version 2 of the
you only look once (YOLO) detector\cite{redmon2016yolo9000}, however
it is compared to Faster RCNN, which is considered by most to be state
of the art. For testing, open-source TensorFlow\cite{tensorflow2015-whitepaper}
versions of the detectors were used\cite{darkflow2017,DBLP:journals/corr/ChenG17a}.

\subsection{Faster RCNN}

\begin{figure}[tbh]
\begin{centering}
\includegraphics[scale=0.4]{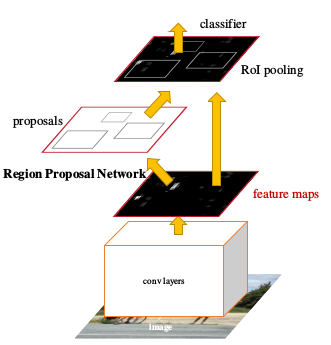}
\par\end{centering}

\caption{\label{fig:Faster-RCNN}Faster RCNN}
\end{figure}

Faster RCNN is the follow on to Fast RCNN\cite{girshickICCV15fastrcnn}
and RCNN\cite{girshick14CVPR}. Faster RCNN starts with a CNN adds
a Region Proposal Network (RPN) to create proposals (bounding boxes)
from the features given by the CNN. Then ROI pooling and a classifier
is used to classify and score each bounding box. A diagram of the
net from the original paper is given in Figure \ref{fig:Faster-RCNN}.
Due to its speed and accuracy, Faster RCNN has been heavily used since
its inception in 2015. 

\subsection{YOLOv2}

YOLOv2 not surprisingly is the improved version YOLO\cite{DBLP:journals/corr/RedmonDGF15},
a net that reached near Faster RCNN accuracy at much greater speeds.
YOLOv2 also starts with a CNN, and then is followed by two more CNNs
that simultaneously produce bounding boxes, object confidence scores,
and class scores. Additionally YOLOv2 includes a route and reorg layers
that allow the net to use features from earlier in the net similar
to SSD\cite{DBLP:journals/corr/LiuAESR15}.

\section{Datasets}

To train a deep neural network object detector requires a vast amount
of data; hence the coupling of deep learning with big data. The neural
nets in the previous section all provide pretrained network weights,
usually started on ImageNet and then on PASCAL VOC\cite{Everingham10}
or MS COCO\cite{DBLP:journals/corr/LinMBHPRDZ14}. Detailed in Table
\ref{tab:Public-Large-Scale}, these datasets provide thousands of
images, and the detectors trained on them are good general purpose
object detectors. However these datasets largely contain imagery taken
from personal cameras at ground level and contain very little to no
aerial data. Additionally these images are relatively low resolution
compared to aerial imagery.

\begin{table}[tbh]
\caption{\label{tab:Public-Large-Scale}Public large scale detection datasets}

\centering{}%
\begin{tabular}{|c|c|}
\hline 
Dataset &
Number of Training Images\tabularnewline
\hline 
\hline 
ImageNet &
450000+\tabularnewline
\hline 
MS COCO &
200000+\tabularnewline
\hline 
PASCAL VOC (2007+2012) &
\textasciitilde{}20000\tabularnewline
\hline 
\end{tabular}
\end{table}

There are several publicly available aerial/elevated imagery detection
datasets. Additionally AFRL has some in house aerial imagery, referred
to as Air Force aerial vehicle imagery dataset (AFVID), that has been
truthed.

\subsection{Dataset descriptions}

\begin{figure*}[tbh]
\begin{centering}
\subfloat[PASCAL VOC]{\includegraphics[width=2.5in]{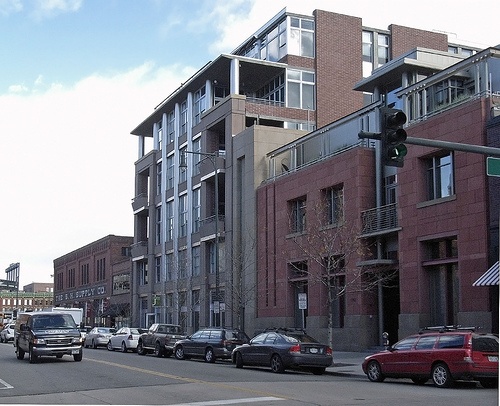}

}\subfloat[VEDAI]{\includegraphics[width=2in]{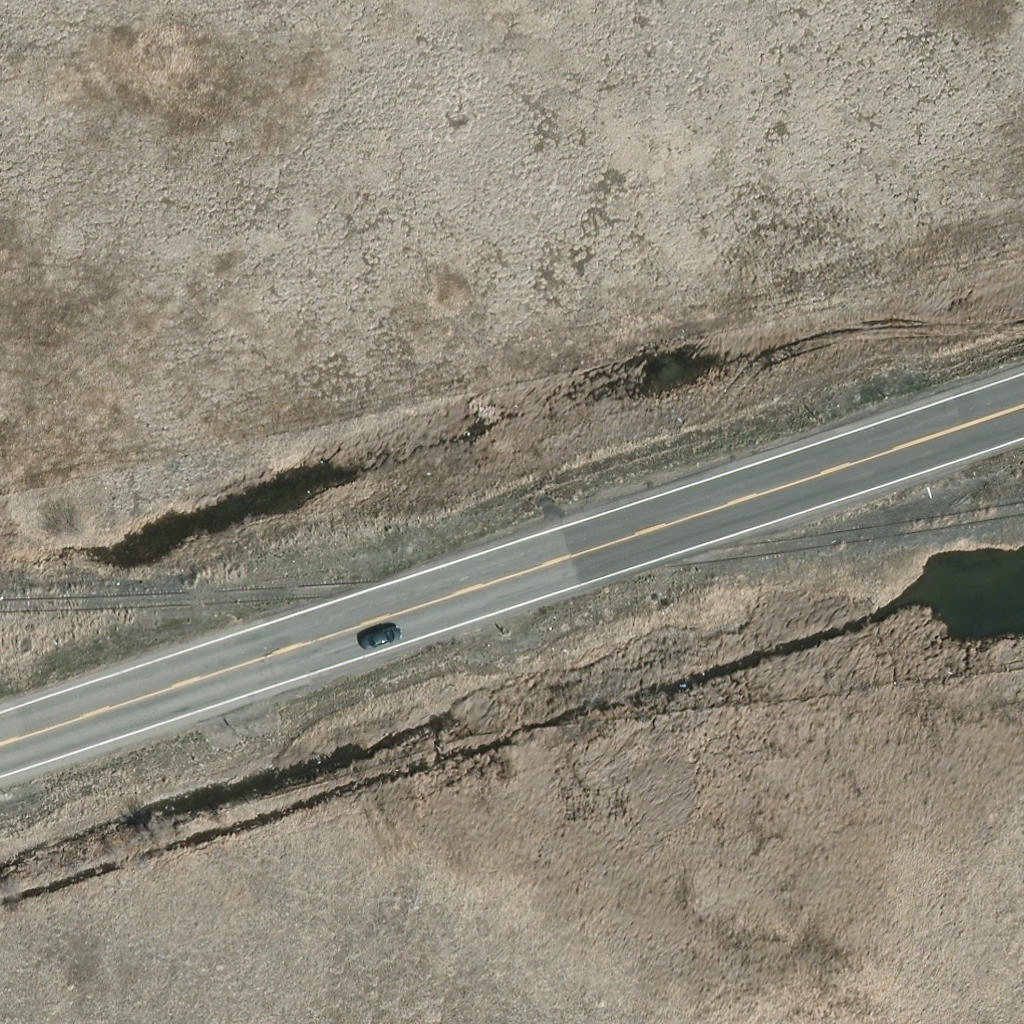}

}
\par\end{centering}

\begin{centering}
\subfloat[AFVID]{\includegraphics[width=2.5in]{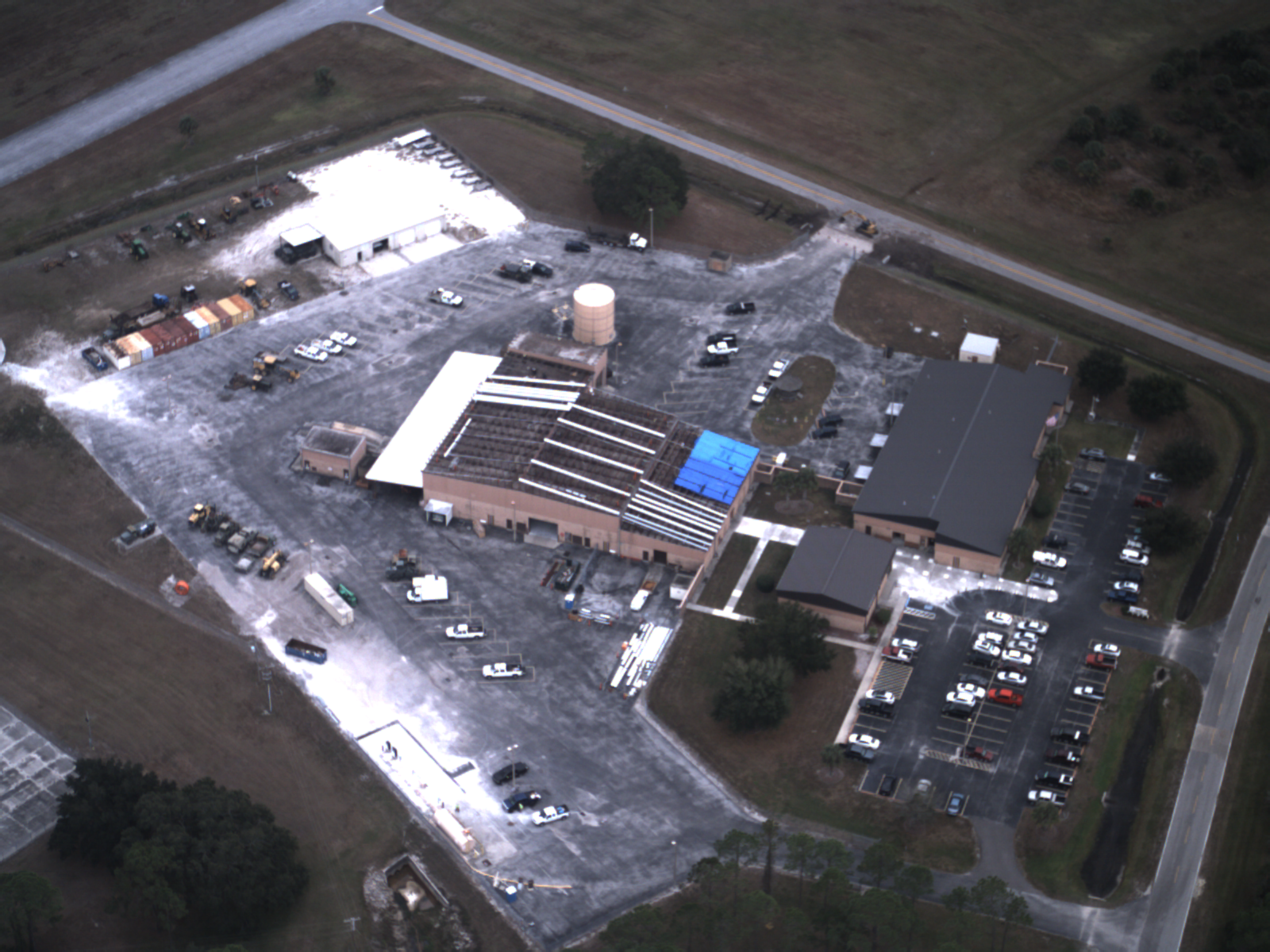}

}\subfloat[AF Building Camera]{\includegraphics[width=2.5in]{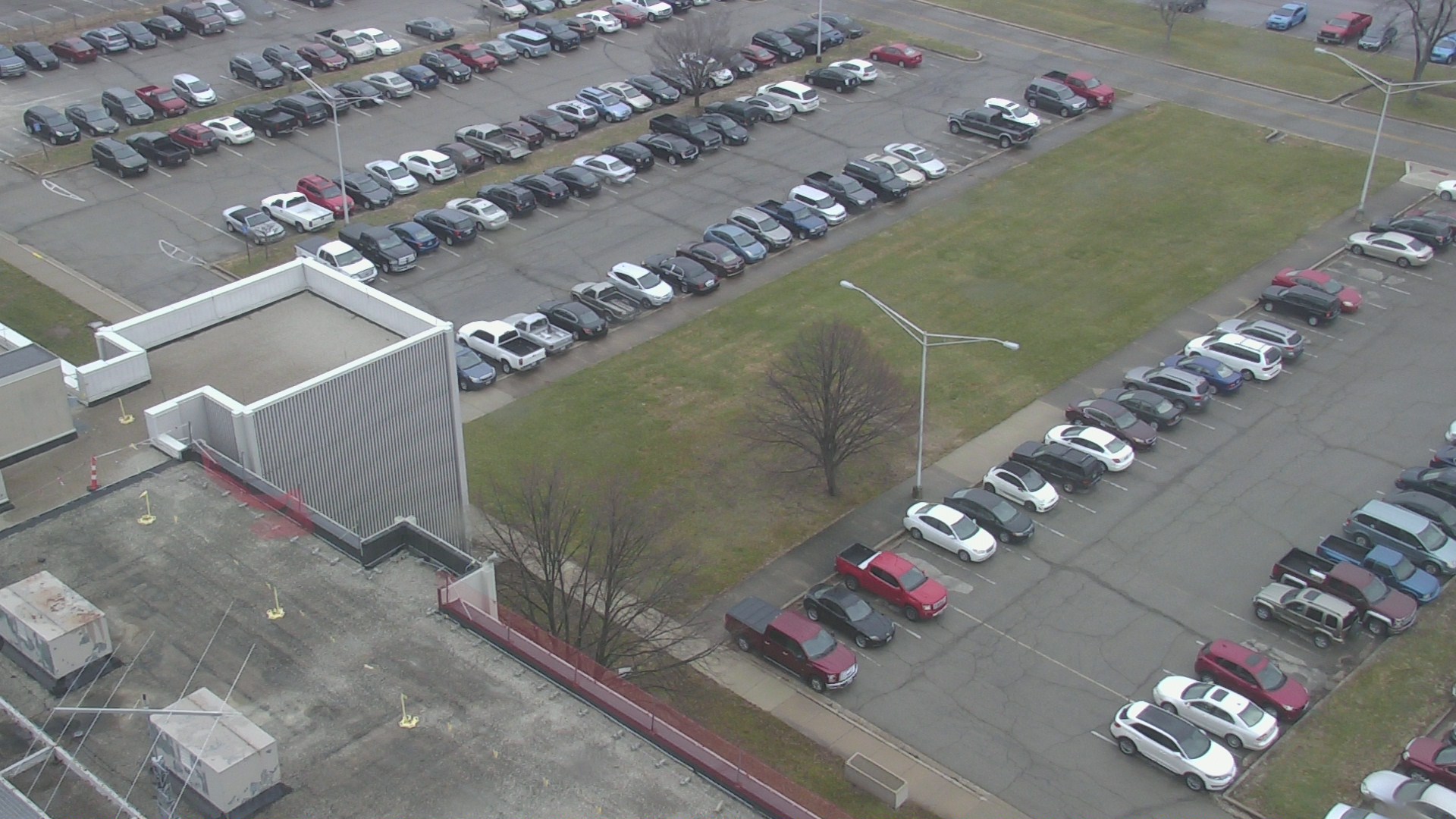}

}
\par\end{centering}

\centering{}\caption{\label{fig:Sample-Imagery}Sample imagery}
\end{figure*}

Aerial data can be surprisingly diverse. It can be from different
view points, have different ground sampling distances (gsd), different
image sizes, aspect ratios, color, etc. Vehicles in different data
may be significantly different in size and appearance. Figure \ref{fig:Sample-Imagery}
shows different images from four different datasets. It is obvious
how different the VOC data is from any of the elevated data, while
the VEDAI and AFVID data are somewhat similarities, and the AF Building
Camera data is more similar to the aerial and satellite data. A more
detailed look at the aerial datasets is available in Table \ref{tab:dataset }. 

\begin{table*}[tbh]
\caption{\label{tab:dataset } Aerial imagery datasets}

\centering{}%
\begin{tabular*}{15cm}{@{\extracolsep{\fill}}|c|>{\centering}p{2.2cm}|>{\raggedright}p{2.55cm}|>{\centering}p{2.5cm}|>{\centering}p{2.6cm}|}
\hline 
Dataset &
Image Width, Height &
\centering{}Mean Vehicle Width,Height &
Ratio Target Area/Image Area &
Percent of Targets Overlapping\tabularnewline
\hline 
\hline 
VEDAI &
1024, 1024 &
\centering{}41.2, 40.8 &
0.0016 &
5\tabularnewline
\hline 
DLR3k &
5616, 3744 &
\centering{}30.4, 30.0 &
0.00004 &
14\tabularnewline
\hline 
NeoVision2-Helicopter &
1920, 1080 &
\centering{}124.6, 85.8 &
0.0069 &
20\tabularnewline
\hline 
AFVID 1 &
1600, 1200 &
\centering{}39.4, 34.2 &
0.0008 &
33\tabularnewline
\hline 
AFVID 2 &
1600, 1200 &
\centering{}55.6, 63.7 &
0.0019 &
54\tabularnewline
\hline 
AFVID 3 &
1600, 1200 &
\centering{}43.5, 38.6 &
0.0009 &
40\tabularnewline
\hline 
AFVID 4 &
1600, 1200 &
\centering{}67.7, 63.5 &
0.0026 &
34\tabularnewline
\hline 
AF Building Camera &
1920, 1080 &
\centering{}51.7, 35.5 &
0.0012 &
53\tabularnewline
\hline 
\end{tabular*}
\end{table*}

\paragraph{VEDAI}

Vehicle Detection in Aerial Imagery (VEDAI) dataset consists of satellite
imagery taken over Utah in 2012. There are 1,210 orthonormal images
in RGB and IR. Only RGB is used here.

\paragraph{DLR3k}

Data from the German Aerospace Center (DLR),there are 20 images taken
from a camera on an airplane about 1000 feet over Munich, Germany.

\paragraph{NeoVision2 - Helicopter}

A set of 32 video clips taken from an HD camera on helicopter flying
in Los Angeles, California area.

\paragraph*{AFVID}

Video clips taken from a small UAV flying about 1000 feet over Avon
Park, Florida or flying about 700 feet over Camp Atterbury, Indiana.

\paragraph{AF Building Camera }

Handful of images taken from cameras on a tower at Wright Patterson
Air Force Base.

\subsection{Dataset segmentation }

Much of aerial imagery is given as video clips, meaning it can not
be considered independent and identically distributed(IID), unlike
the the large detection image datasets. This means even though there
may be thousands of images, some of it may be so close in appearance
that effectively there is less data and none of our datasets are anywhere
near the size of even PASCAL VOC. Ideally each dataset should be separated
into train (60\%), validation (20\%), and test (20\%) sets. Typically
this should only be done with datasets that have independent images.
Some of the datasets consist of a short video sequences where at least
two videos are kept out, one for validation and one for testing.

\section{Modifying YOLOv2}

YOLO provides models and pretrained weights for the MS COCO and PASCAL
VOC datasets, both which are capable of detecting vehicles that are
relatively large in the image. To detect objects that are small relative
to the size of the image and to detect different numbers of classes
the YOLO net needs to be modified and fine-tuned using appropriate
data.

\subsection{Number of classes}

In the configuration file that defines the YOLO net model there is
a 'classes' definition that is used to define the number of classes.
To change the number of classes used, more the just the classes setting
will need to be changed; the number of filters in the last convolutional
layer must be altered to reflect the changed number of classes. The
number of filters is set by $num(classes+coords+1)$, where num is
number of anchor boxes, and coords is four corresponding to the four
coordinates used to define a bounding box.

\subsection{Net resolution and depth\label{sub:Net-Resolution-and}}

The standard YOLOv2 net has input resolution of 416x416 and after
pass-through and max-pooling has an output feature resolution of 26x26
and 13x13. For a 1024x1024 image that would create feature resolutions
corresponding to about 40 and 80 pixels, which is approximately the
size of and twice the size of the vehicles in the VEDAI imagery, meaning
a vehicle may correspond to a single point on a feature map. Ideally
there should be multiple points per vehicle. To increase net feature
resolution and consequently the number of feature points per vehicle,
there are two methods: increase the input resolution or to decrease
the net depth. 

Increasing the net input resolution simply means increasing the width
and height of the first layer of the net. In the previous example
doubling the width and height leads to output sizes of 52x52 and 26x26
and 20/40 pixels per feature, so that there can be about four feature
points for the average vehicle. However, this significantly increases
GPU memory usage and decreases the speed of the net. 

To decrease the net depth, convolutional and max pooling layers are
removed so the net is downsampled less, making the net shallower.
This goes against the now conventional wisdom that deeper nets are
typically better, but has been shown to work well on aerial data\cite{HusterMSS,sakla2017}.
Removing one max pooling layer and associated CNN layers gives the
same output resolution as doubling the input resolution without having
as great as an effect the memory usage or speed of the net. A sample
shallower YOLO net is given in Appendix A Table \ref{tab:Shallower-YOLOv2}
alongside a typical net.

\subsection{Net shape }

YOLO provide pretrained square shaped nets, however most of our data
is not square. In YOLOv2, the shape of the net can be changed to closer
match the aspect ratio of the input data. Faster RCNN does this automatically.

\subsection{Anchors}

In \cite{sakla2017}, the Faster RCNN anchors were fixed, as the bounding
boxes were refined to be of fixed shape and size. For multi-scale
detection, it should be better to have multiple anchor box sizes.
The five anchor sizes in YOLOv2 were determined from a k-means approach
using the bounding boxes in the VOC dataset. Due to the difference
in orientation and scale of vehicles, most of the anchors were kept,
with only the largest being removed, since it is not expected that
there will be large objects in the image.

\section{Experiments and results}

To calculate how the detectors are performing, two metrics are used: 

\begin{equation}
precision=tp/(fp+tp)
\end{equation}
and 

\begin{equation}
recall=tp/nObjects.
\end{equation}
In defense applications, the false alarm rate (FAR) is often used
instead of precision but 
\begin{equation}
FAR=1-precision,
\end{equation}
and recall is sometimes referred to the detection rate. The another
metric used here is frames per second (FPS). To match detections to
ground truth, intersection over union (IOU) is used. IOU is ratio
between the area in which the two boxes overlap and the total area
of each box not including the overlap. 

A detector was desired for aerial imagery with small vehicles. Only
a single class, vehicle, was used in training and testing. Therefore
average precision (AP) and average recall (AR) refer to the average
over all the test images at an IOU threshold of 0.5.

\subsection{Pretrained object detectors}

Most open source detectors provide their best trained for PASCAL VOC
and/or MS COCO. Since these datasets are large and somewhat diverse
they make good general purpose object detectors. The results for these
detectors tested on aerial imagery is given in Table \ref{tab:Pretrained-Object-Detectors}.
While Faster RCNN gives the best precision and recall, it is the slowest
of the tested detectors. All of the detectors performed quite poorly
without fine-tuning for aerial data. 

\begin{table*}[tbh]
\caption{\label{tab:Pretrained-Object-Detectors}Pretrained object detectors
tested on aerial data}

\centering{}%
\begin{tabular}{|c|c|c|c|c|}
\hline 
Net - trained &
AFVID &
VEDAI &
AF Building Camera &
FPS\tabularnewline
\hline 
\hline 
YOLO - VOC+COCO &
AP: 0.0 AR: 0.03 &
AP: 0.0 AR: 0.03 &
AP: 0.05 AR: 0.07 &
16 \tabularnewline
\hline 
YOLO - VOC  &
AP: 0.0 AR: 0.0 &
AP: 0.0 AR: 0.01  &
AP: 0.04 AR: 0.04 &
15 \tabularnewline
\hline 
Faster RCNN - COCO &
AP: 0.04 AR: 0.18 &
AP: 0.01 AR: 0.12 &
AP: 0.13 AR: 0.19 &
3\tabularnewline
\hline 
\end{tabular}
\end{table*}

\subsection{YOLO modifications }

The biggest increases in performance come from altering the net's
size and depth. Table \ref{tab:Results---Fine-tuned} compares the
performance of results of several modified YOLO nets and a shallow
Faster RCNN net based on \cite{sakla2017}. All the nets were fine-tuned
using AFVID and VEDAI data. The YOLO nets were based off of the YOLO-VOC
net but Faster RCNN was from COCO. The first YOLO listed is the standard
YOLO modified for one class (vehicles). While it is five times faster
than Faster RCNN, the performance across the board is much weaker.
Doubling the size as described in \ref{sub:Net-Resolution-and}, gives
a performance boost, particularly on the AFVID dataset, but halves
the speed. Removing several convolutional layers and a max pooling
layer greatly improves precision, recall, and speed. With the standard
input resolution but shallower, YOLO is approximately seven times
faster the shallower Faster RCNN, but still has lower recall and lower
precision on the AFVID and VEDAI datasets. Increasing the size of
the shallower net decreases the speed of the net again, but still
boosts precision and recall. Changing the input shape of the shallow
YOLO net to better match the aspect ratio of the AFVID data causes
a large increase in the precision for that data, but causes a small
decrease on the square VEDAI data. On the AF Building Camera data,
which none of the nets were trained on, the shallow YOLO nets do better
than Faster RCNN. 

\begin{table*}[tbh]
\caption{\label{tab:Results---Fine-tuned}Results fine-tuned on AFVID and VEDAI
data}

\centering{}%
\begin{tabular}{|c|c|c|c|c|c|}
\hline 
Net  &
Input Size &
AFVID &
VEDAI &
AF Building Camera &
FPS\tabularnewline
\hline 
\hline 
Shallow Faster RCNN  &
600xA &
AP: 0.88 AR: 0.89 &
AP: 0.78 AR: 0.91 &
AP: 0.05 AR: 0.08 &
3\tabularnewline
\hline 
YOLO &
416x416 &
AP: 0.0 AR: 0.03 &
AP: 0.0 AR: 0.01 &
AP: 0.01 AR: 0.02 &
15\tabularnewline
\hline 
YOLO &
832x832 &
AP: 0.01 AR: 0.13 &
AP: 0.0 AR: 0.05 &
AP: 0.0 AR: 0.04 &
6\tabularnewline
\hline 
Shallow YOLO &
416x416 &
AP: 0.1 AR: 0.37 &
AP: 0.46 AR: 0.72 &
AP: 0.14 AR: 0.21 &
25\tabularnewline
\hline 
Shallow YOLO &
832x832 &
AP: 0.21 AR: 0.79 &
AP: 0.66 AR: 0.91 &
AP: 0.15 AR: 0.3 &
8\tabularnewline
\hline 
Shallow YOLO &
864x480 &
AP: 0.76 AR: 0.87 &
AP: 0.61 AR: 0.81 &
AP: 0.21 AR: 0.3 &
13\tabularnewline
\hline 
\end{tabular}
\end{table*}
One application is to use this CNN network on live data from the AF
Building Cameras, which has a larger variance in vehicle sizes than
most aerial datasets. The nets were trained on the AFVID data and
a few images from the AF Building Cameras. The results in Table \ref{tab:Cerb+Aerial},
show that rectangular shallow YOLO outperforms Faster RCNN on this
data, while the shallow Faster RCNN is still better on AFVID and VEDAI
data.

\begin{table*}[tbh]
\caption{\label{tab:Cerb+Aerial}Results fine-tuned on AFVID and AF Building
Camera data}

\centering{}%
\begin{tabular}{|c|c|c|c|c|c|}
\hline 
Net &
Input Size &
AFVID &
VEDAI &
AF Building Camera &
FPS\tabularnewline
\hline 
\hline 
Shallow Faster RCNN &
600xA &
AP: 0.88 AR: 0.89 &
AP: 0.1 AR: 0.46 &
AP: 0.42 AR: 0.58 &
3\tabularnewline
\hline 
Shallow YOLO  &
416x416 &
AP: 0.13 AR: 0.47 &
AP: 0.03 AR: 0.22 &
AP: 0.25 AR: 0.39 &
25\tabularnewline
\hline 
Shallow YOLO  &
864x480 &
AP: 0.74 AR: 0.82 &
AP: 0.05 AR: 0.27 &
AP: 0.5 AR: 0.66 &
13\tabularnewline
\hline 
\end{tabular}
\end{table*}

\subsection{Deeper look at results}

For cases where the input to the net is an entire image, the object
size relative the size of image is more important to than the size
of the object itself. For example, in the MSCOCO dataset images are
typically 640x480 and the benchmark define a small object to be less
than 32x32 a medium object between 32x32 and 96x96. While most of
the mean object sizes in the aerial imagery fall in the medium category
by this definition, the image sizes are larger so when resized to
the input resolution of the nets they equivalently fall into the small
category. Hence the categories were redefined in terms of the image
size, such that small objects are 0.1\% or less, medium objects are
between 0.1\% and 0.3\% and large objects are greater than 0.3\% of
the original image. Looking at the results in terms of size on AF
Building Camera data in Table \ref{tab:Cerberus-Camera-Object} shows
a surprising result for Faster RCNN, the medium size is actually has
the worst performance. Shallow YOLO performs as expected.

\begin{table*}[tbh]
\caption{\label{tab:Cerberus-Camera-Object}AF building camera object size}

\centering{}%
\begin{tabular}{|c|c|c|c|c|}
\hline 
Detector &
mAP &
mAP Large &
mAP Medium &
mAP Small\tabularnewline
\hline 
\hline 
Shallow Faster RCNN &
0.080 &
0.171 &
0.080 &
0.117\tabularnewline
\hline 
Shallow YOLO (864x480) &
0.124 &
0.404 &
0.329 &
0.102\tabularnewline
\hline 
\end{tabular}
\end{table*}

The irregularity in the medium size objects in Faster RCNN may be
due to a quirk in the small AF Building Camera dataset. Shown in Figure
3 is a set images taken by the same camera at different times, with
the last two sets of images taken only a few minutes apart. Both detectors
do well when the cars are sparse and most have visible separation.
However when the parking lot is full and the cars are densely packed
the Faster RCNN sometimes fails to detect some of the vehicles while
YOLO does a much better job, irregardless of illumination. Therefore
the poor performance on medium sized objects by this Faster RCNN net
may be due to its inability handle heavily overlapping objects in
some images.

\begin{figure*}[tbh]
\begin{centering}
\subfloat[Faster RCNN Low Density]{\includegraphics[scale=0.1]{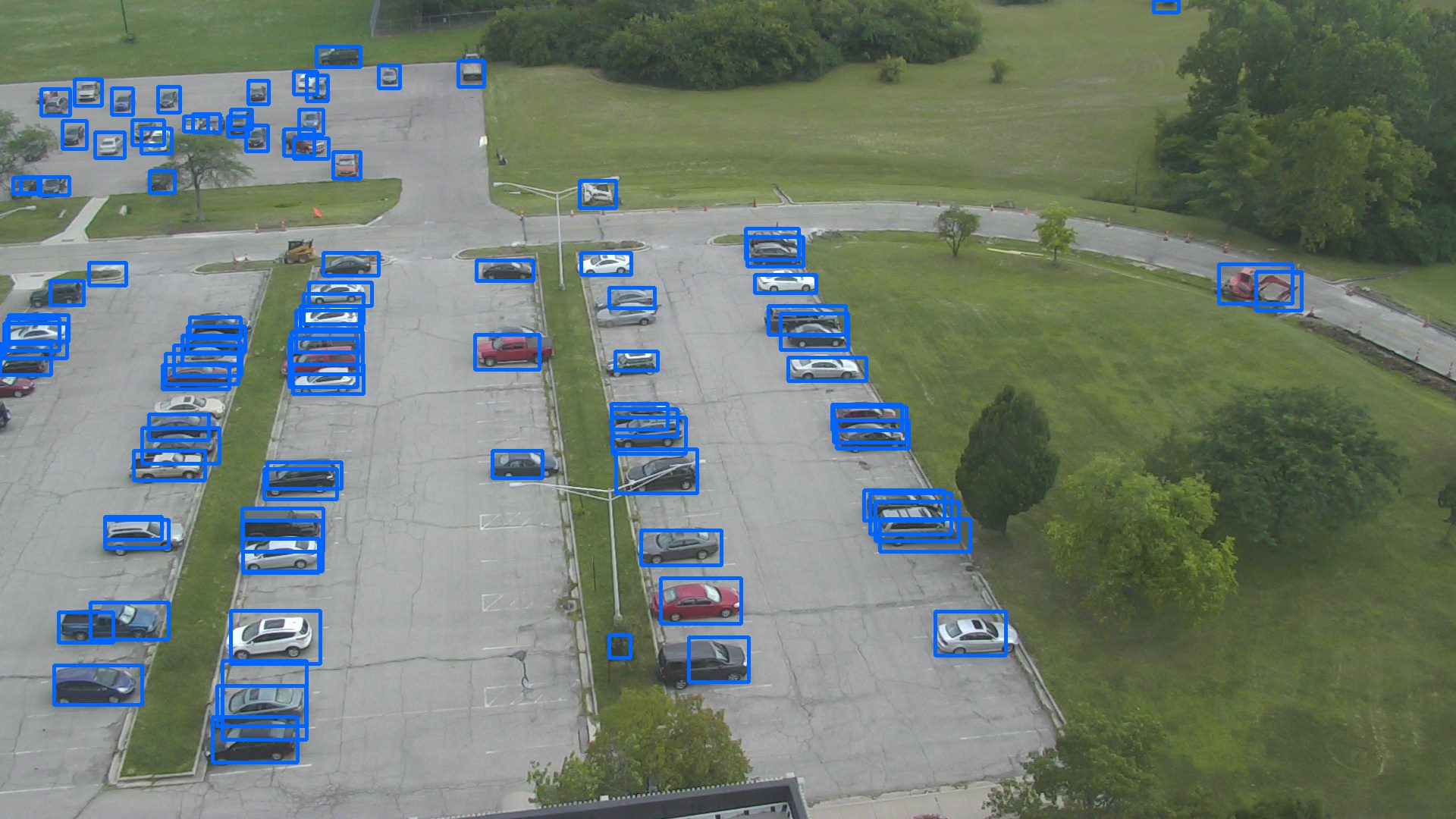}

}\subfloat[YOLO Low Density]{\includegraphics[scale=0.1]{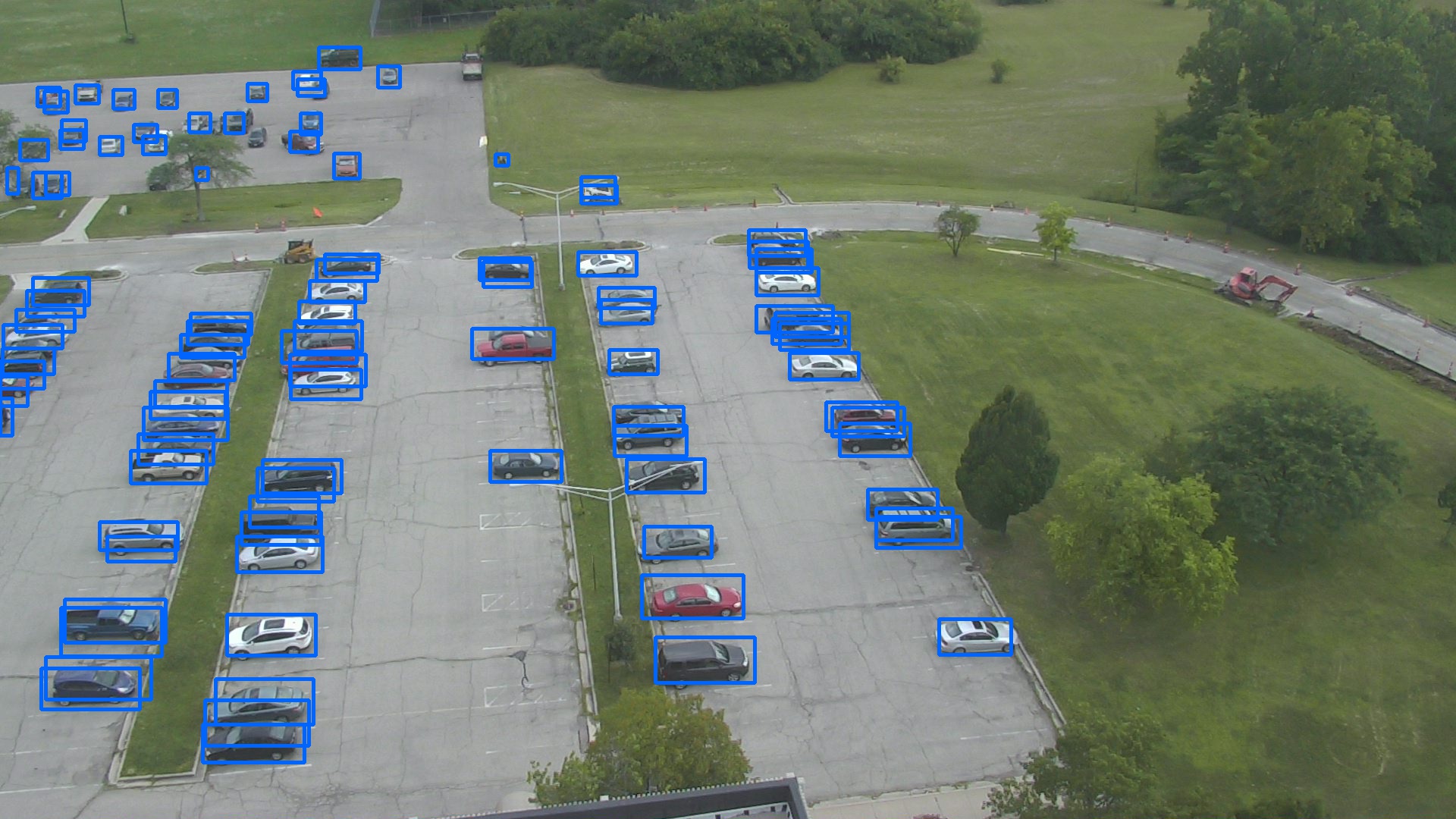}

}
\par\end{centering}

\begin{centering}
\subfloat[Faster RCNN High Density, High Illumination]{\includegraphics[scale=0.1]{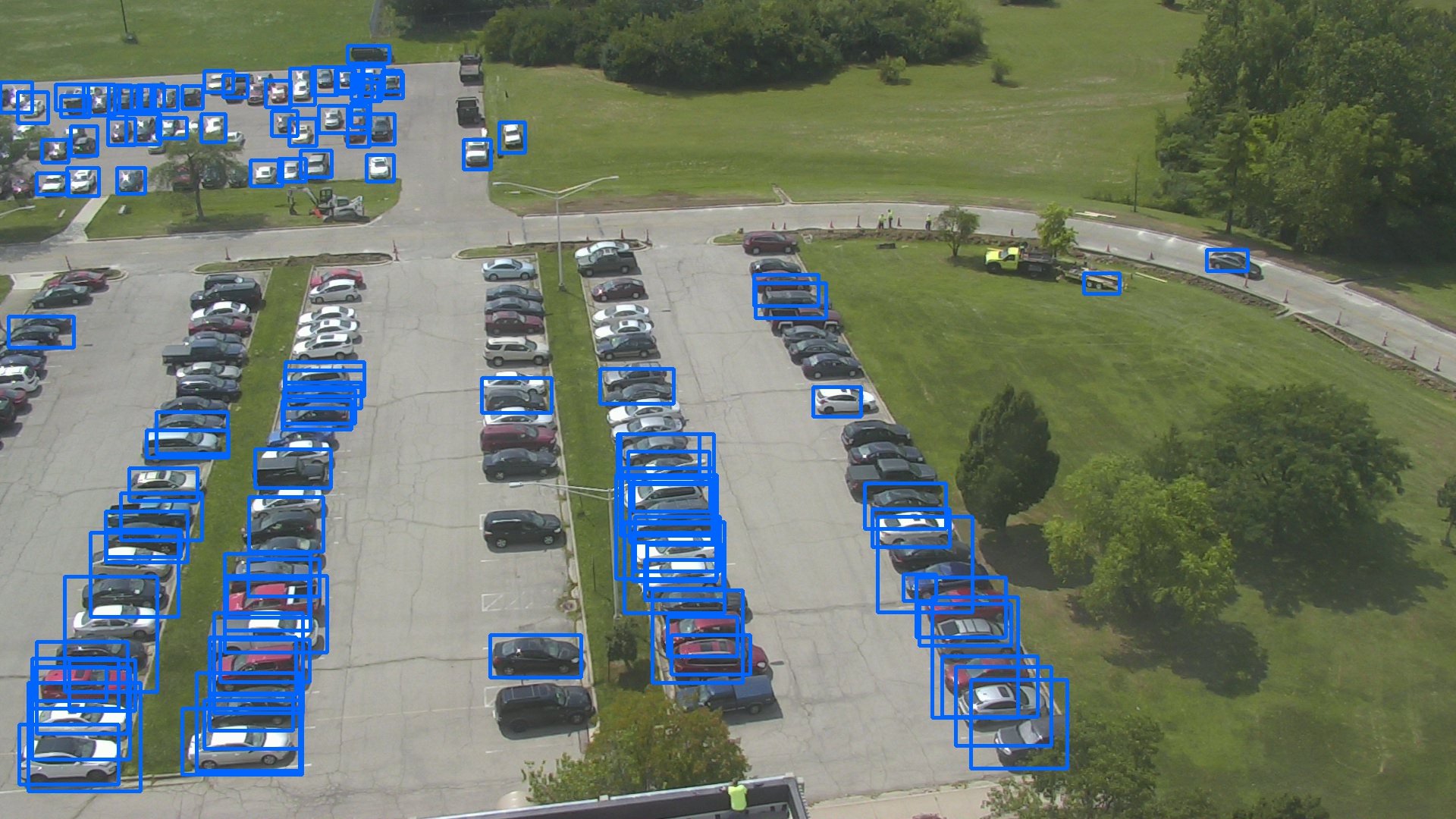}

}\subfloat[YOLO High Density, High Illumination]{\includegraphics[scale=0.1]{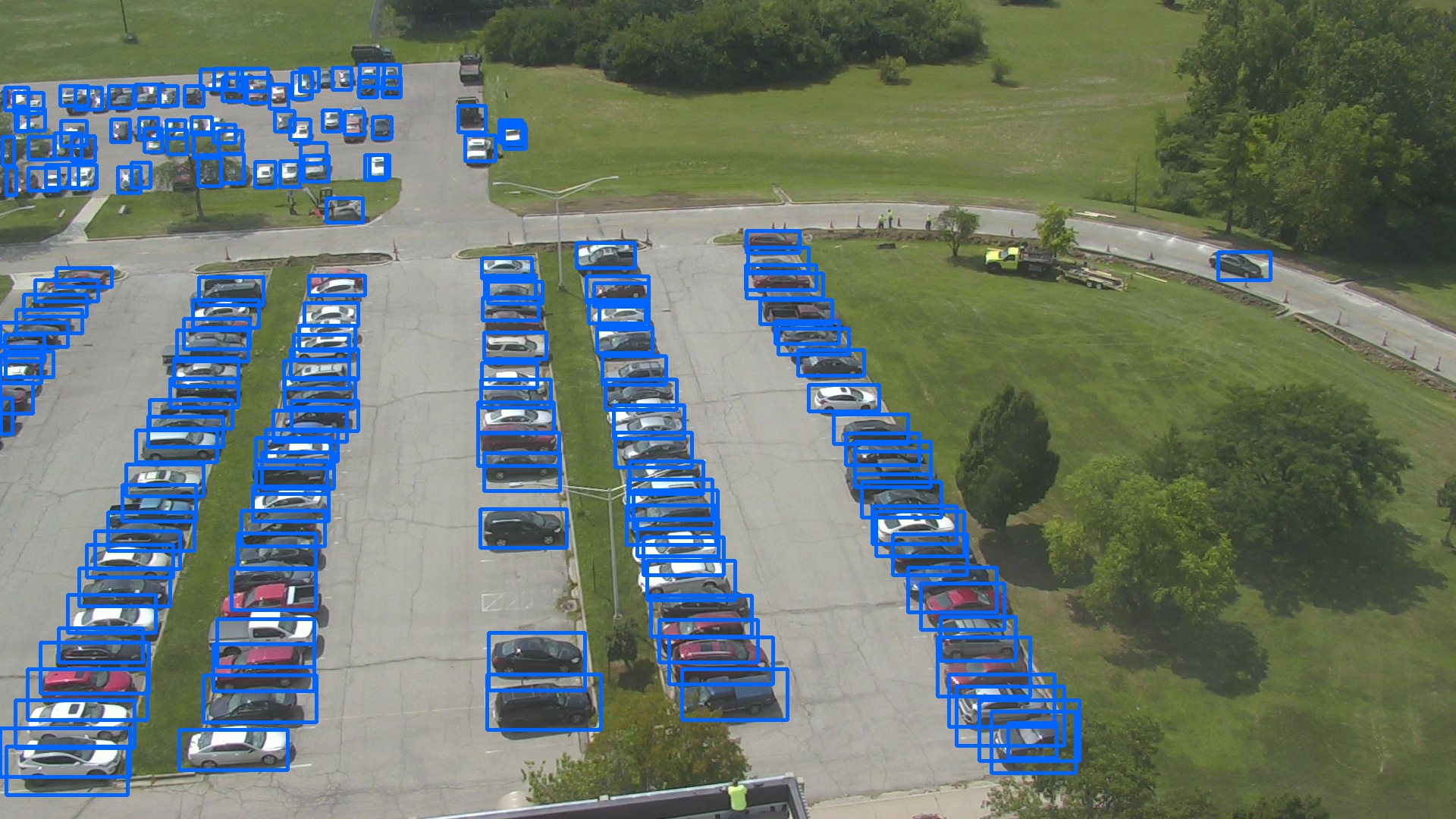}

}
\par\end{centering}

\begin{centering}
\subfloat[Faster RCNN High Density, Low Illumination]{\includegraphics[scale=0.1]{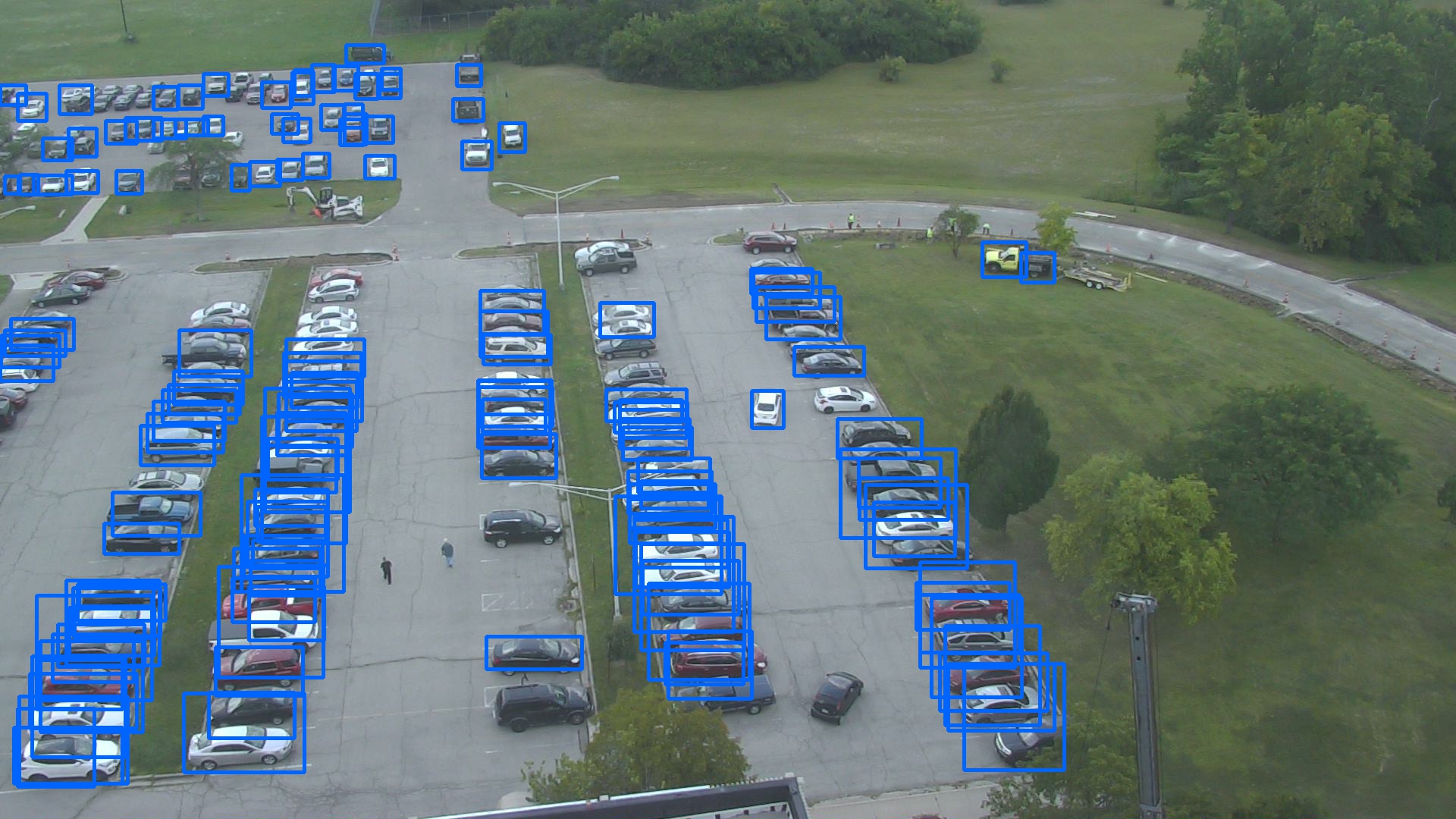}

}\subfloat[YOLO High Density, Low Illumination]{\includegraphics[scale=0.1]{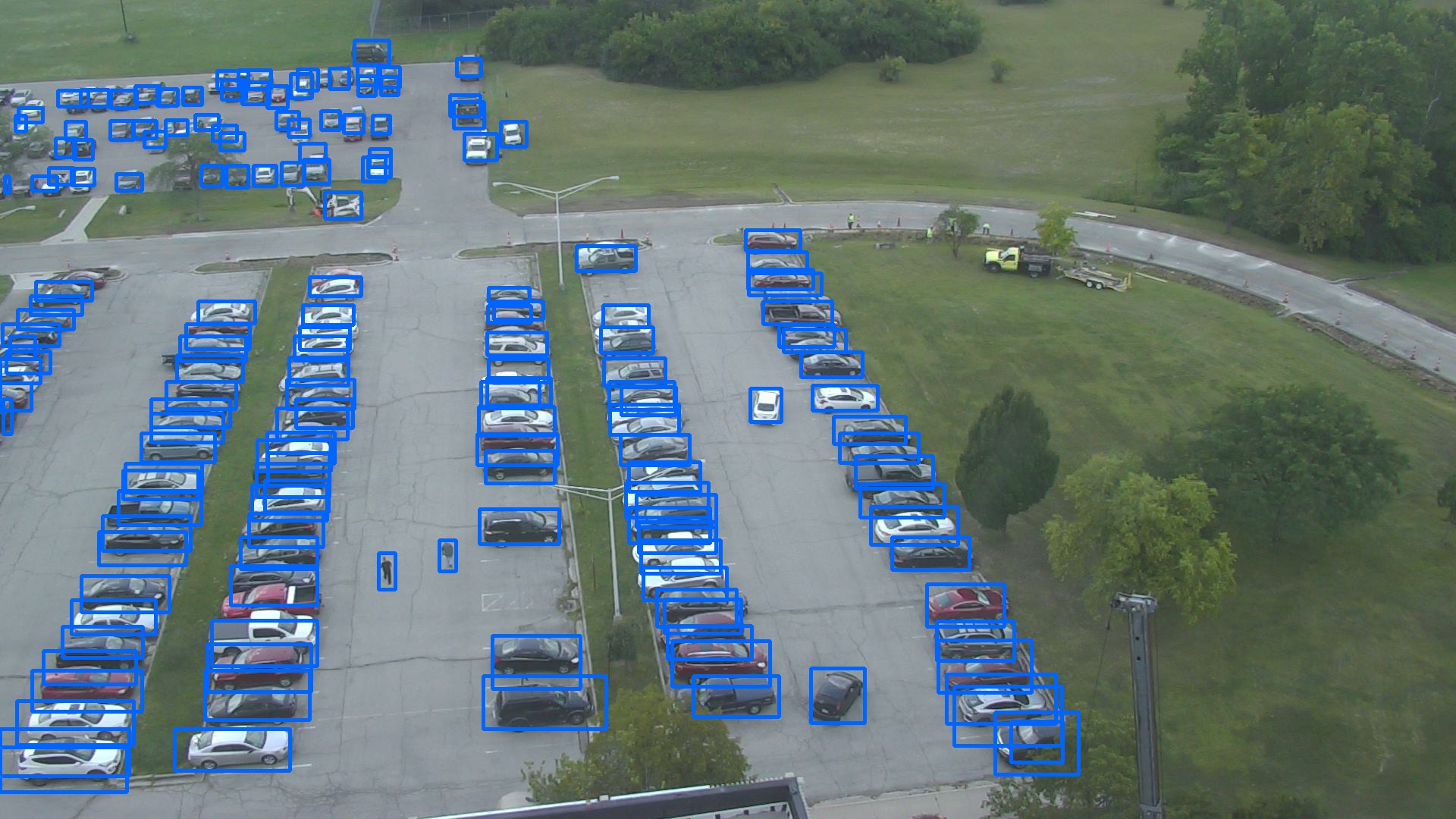}}
\par\end{centering}

\caption{Object density and illumination}
\end{figure*}

\section{Conclusions}

While the out of the box YOLO performs poorly on aerial imagery, a
few modifications can greatly improve its performance. First making
the net shallower to increase its output resolution. Second changing
the net shape to more closer match the aspect ratio of the data. While
the modified YOLO's precision and recall is still typically a bit
worse than Faster RCNN, it's increased speed makes a it good option
when near real-time vehicle detectors for aerial imagery are required. 

\bibliographystyle{12_home_macer_Documents_paper_ieee}
\phantomsection\addcontentsline{toc}{section}{\refname}\bibliography{11_home_macer_Documents_paper_reference,IEEEabrv,IEEEexample}

\appendix

\section*{Appendix}

\begin{table*}[tbh]
\caption{YOLOv2 nets}
\subfloat[Standard YOLOv2 net]{

\begin{tabular}{|c|c|c|c|}
\hline 
layer &
filters &
kernal size &
net size\tabularnewline
\hline 
\hline 
Input &
- &
- &
416 x 416 x 3\tabularnewline
\hline 
Convolutional &
32 &
3 x 3 / 1 &
416 x 416 x 32\tabularnewline
\hline 
Max Pooling &
- &
2 x 2 / 2 &
208 x 208 x 32\tabularnewline
\hline 
Convolutional &
64 &
3 x 3 / 1 &
208 x 208 x 64\tabularnewline
\hline 
Max Pooling &
- &
2 x 2 / 2 &
104 x 104 x 64\tabularnewline
\hline 
Convolutional &
128 &
3 x 3 / 1 &
104 x 104 x 128\tabularnewline
\hline 
Convolutional &
64 &
1 x 1 / 1 &
104 x 104 x 64\tabularnewline
\hline 
Convolutional &
128 &
3 x 3 / 1 &
104 x 104 x 128\tabularnewline
\hline 
Max Pooling &
- &
2 x 2 / 2 &
52 x 52 x 128\tabularnewline
\hline 
Convolutional &
256 &
3 x 3 / 1 &
52 x 52 x 256\tabularnewline
\hline 
Convolutional &
128 &
1 x 1 / 1 &
52 x 52 x 128\tabularnewline
\hline 
Convolutional &
256 &
3 x 3 / 1 &
52 x 52 x 256\tabularnewline
\hline 
Max Pooling &
- &
2 x 2 / 2 &
26 x 26 x 256\tabularnewline
\hline 
Convolutional &
512 &
3 x 3 / 1 &
26 x 26 x 512\tabularnewline
\hline 
Convolutional &
256 &
1 x 1 / 1 &
26 x 26 x 256\tabularnewline
\hline 
Convolutional &
512 &
3 x 3 / 1 &
26 x 26 x 512\tabularnewline
\hline 
Convolutional &
256 &
1 x 1 / 1 &
26 x 26 x 256\tabularnewline
\hline 
Convolutional &
512 &
3 x 3 / 1 &
26 x 26 x 512\tabularnewline
\hline 
Max Pooling &
- &
2 x 2 / 2 &
13 x 13 x 512\tabularnewline
\hline 
Convolutional &
1024 &
3 x 3 / 1 &
13 x 13 x 1024\tabularnewline
\hline 
Convolutional &
512 &
1 x 1 / 1 &
13 x 13 x 512\tabularnewline
\hline 
Convolutional &
1024 &
3 x 3 / 1 &
13 x 13 x 1024\tabularnewline
\hline 
Convolutional &
1024 &
1 x 1 / 1 &
13 x 13 x 1024\tabularnewline
\hline 
Convolutional &
1024 &
3 x 3 / 1 &
13 x 13 x 1024\tabularnewline
\hline 
Convolutional &
1024 &
3 x 3 / 1 &
13 x 13 x 1024\tabularnewline
\hline 
Convolutional &
1024 &
3 x 3 / 1 &
13 x 13 x 1024\tabularnewline
\hline 
Route &
- &
- &
26 x 26 x 512\tabularnewline
\hline 
Convolutional &
64 &
1 x 1 / 1 &
26 x 26 x 64\tabularnewline
\hline 
Reorg &
- &
- &
13 x 13 x 256\tabularnewline
\hline 
Route &
- &
- &
13 x 13 x 1280\tabularnewline
\hline 
Convolutional &
1024 &
3 x 3 / 1 &
13 x 13 x 1024\tabularnewline
\hline 
Convolutional &
125 &
1 x 1 / 1 &
13 x 13 x 125\tabularnewline
\hline 
\end{tabular}}\subfloat[\label{tab:Shallower-YOLOv2}Shallow YOLOv2 net]{

\begin{tabular}{|c|c|c|c|}
\hline 
layer &
filters &
kernal size &
net size\tabularnewline
\hline 
\hline 
Input &
- &
- &
416 x 416 x 3\tabularnewline
\hline 
Convolutional &
32 &
3 x 3 / 1 &
416 x 416 x 32\tabularnewline
\hline 
Max Pooling &
- &
2 x 2 / 2 &
208 x 208 x 32\tabularnewline
\hline 
Convolutional &
64 &
3 x 3 / 1 &
208 x 208 x 64\tabularnewline
\hline 
Max Pooling &
- &
2 x 2 / 2 &
104 x 104 x 64\tabularnewline
\hline 
Convolutional &
128 &
3 x 3 / 1 &
104 x 104 x 128\tabularnewline
\hline 
Convolutional &
64 &
1 x 1 / 1 &
104 x 104 x 64\tabularnewline
\hline 
Convolutional &
128 &
3 x 3 / 1 &
104 x 104 x 128\tabularnewline
\hline 
Max Pooling &
- &
2 x 2 / 2 &
52 x 52 x 128\tabularnewline
\hline 
Convolutional &
256 &
3 x 3 / 1 &
52 x 52 x 256\tabularnewline
\hline 
Convolutional &
128 &
1 x 1 / 1 &
52 x 52 x 128\tabularnewline
\hline 
Convolutional &
256 &
3 x 3 / 1 &
52 x 52 x 256\tabularnewline
\hline 
Max Pooling &
- &
2 x 2 / 2 &
26 x 26 x 256\tabularnewline
\hline 
Convolutional &
512 &
3 x 3 / 1 &
26 x 26 x 512\tabularnewline
\hline 
Convolutional &
256 &
1 x 1 / 1 &
26 x 26 x 256\tabularnewline
\hline 
Convolutional &
512 &
3 x 3 / 1 &
26 x 26 x 512\tabularnewline
\hline 
Convolutional &
256 &
1 x 1 / 1 &
26 x 26 x 256\tabularnewline
\hline 
Convolutional &
512 &
3 x 3 / 1 &
26 x 26 x 512\tabularnewline
\hline 
Route &
- &
- &
52 x 52 x 256\tabularnewline
\hline 
Convolutional &
64 &
1 x 1 / 1 &
52 x 52 x 64\tabularnewline
\hline 
Reorg &
- &
- &
26 x 26 x 256\tabularnewline
\hline 
Route &
- &
- &
26 x 26 x 768\tabularnewline
\hline 
Convolutional &
1024 &
3 x 3 / 1 &
26 x 26 x 1024\tabularnewline
\hline 
Convolutional &
125 &
1 x 1 / 1 &
26 x 26 x 30\tabularnewline
\hline 
\end{tabular}}
\end{table*}

\end{document}